\documentclass[runningheads]{llncs}

\usepackage[T1]{fontenc}
\usepackage{graphicx}
\usepackage{booktabs}
\usepackage{amsmath}
\usepackage{xcolor}
\usepackage{multirow}
\usepackage{diagbox}

\begin{document}
\title{Edge-Based Learning for Improved Classification Under Adversarial Noise}

\author{Manish Kansana\inst{1} \and
Keyan Alexander Rahimi\inst{2}\and
Elias Hossain\inst{1} \and
Iman Dehzangi\inst{3,4,5}\and 
Noorbakhsh Amiri Golilarz\inst{6}
}
\authorrunning{F. Author et al.}

\author{Manish Kansana\inst{1} \and
Keyan Alexander Rahimi\inst{2} \and
Elias Hossain\inst{1} \and 
Iman Dehzangi\inst{3,4,5} \and
Noorbakhsh Amiri Golilarz\inst{6}
}

\authorrunning{Manish Kansana et al.}

\institute{Department of Computer Science and Engineering, Mississippi State University, Mississippi State, MS 39762, USA\\
\email{mk1684, mh3511}@msstate.edu \and
Brown University, Providence, RI 02912, USA\\
\email{keyan\_rahimi@brown.edu} \and 
Center for Computational and Integrative Biology, Rutgers University, Camden, NJ, USA \and 
Department of Computer Science, Rutgers University, Camden, NJ, USA \and 
Rutgers Cancer Institute, Rutgers University, New Brunswick, NJ 08901, USA\\
\email{i.dehzangi@rutgers.edu}
\and 
Department of Computer Science, The University of Alabama, Tuscaloosa, AL 35487, USA\\
\email{noor.amiri@ua.edu} 
}

\maketitle              % typeset the header of the contribution
\begin{abstract}
Adversarial noise introduces small perturbations in images, misleading deep learning models into misclassification and significantly impacting recognition accuracy. In this study, we analyzed the effects of Fast Gradient Sign Method (FGSM) adversarial noise on image classification and investigated whether training on specific image features can improve robustness. We hypothesize that while adversarial noise perturbs various regions of an image, edges may remain relatively stable and provide essential structural information for classification.
To test this, we conducted a series of experiments using brain tumor and COVID datasets. Initially, we trained the models on clean images and then introduced subtle adversarial perturbations, which caused deep learning models to significantly misclassify the images. Retraining on a combination of clean and noisy images led to improved performance. To evaluate the robustness of the edge features, we extracted edges from the original/clean images and trained the models exclusively on edge-based representations. When noise was introduced to the images, the edge-based models demonstrated greater resilience to adversarial attacks compared to those trained on the original or clean images. These results suggest that while adversarial noise is able to exploit complex non-edge regions significantly more than edges, the improvement in the accuracy after retraining is marginally more in the original data as compared to the edges. Thus, leveraging edge-based learning can improve the resilience of deep learning models against adversarial perturbations.

\end{abstract}

\section{Introduction}

Machine learning models have evolved rapidly in the past few years, from basic systems to integral components across a myriad of applications, transforming industries and societal functions. At its core, it enables systems to learn from data, identify patterns, and make decisions with minimal human intervention. This capability has been harnessed to tackle complex problems, optimize processes, and provide innovative solutions in diverse fields. As technology advances and data proliferates, the importance and influence of machine learning models, especially in image classification, continues to expand, making them indispensable tools in our increasingly data-driven world. These vision models are already applied in many different vital industries ranging from autonomous driving to healthcare and security. All of these areas require a high amount of accuracy, as a mistake in classification can be dangerous at best, and fatal at worst. That is why it is imperative that these image classification models be robust and powerful enough to avoid errors, both accidental and purposeful. 

An adversarial attack is a term used to describe a system designed to fool machine learning models into making mistakes, and in our case, misclassifying an image. They exploit weaknesses in the models and take advantage of properties such as short-cutting in order to subtly alter an image just enough to make the model classify an image differently. Modern machine learning models are susceptible to these attacks, as they have inherent limitations and vulnerabilities, ranging from the data they are trained on to their parameters and lack of transparency \cite{grosse2017statistical}. Through these multiple limitations, adversarial attacks can be created with surprising ease, and go on to become very effective and discreet, as they are often not detectable by simply looking at the images \cite{zhang2022adversarial}. Attack methods such as PGD \cite{geisler2024attacking} and FGSM \cite{goodfellow2014explaining} are examples of this. Even high performing state-of-the-art models can be very susceptible to attack \cite{zhang2018efficient}. There have been many different strategies and methods implemented to counter adversarial attacks. It is extremely important to study adversarial noise removal and work to eliminate malicious attempts, as successful attacks could lead to disastrous consequences in many fields, such as autonomous driving or in healthcare \cite{akhtar2018threat}. 

There are currently multiple existing methods developed in order to tackle the issue of adversarial noise, which range in terms of complexity and tactic \cite{ayas2022projected}, \cite{huang2017adversarial}, \cite{kurakin2018adversarial}. However, these defenses often fall shorts of expectation. There is currently a great need for more robust systems to serve as a defense against adversarial noise, as machine learning only becomes increasingly important in numerous industries where high accuracy is a necessity rather than a luxury \cite{carlini2017towards}. By mitigating these attacks, we open up many more possibilities for neural networks to be effective in more areas.

Current defenses against adversarial attacks include many different strategies, such as detection-based methods, domain area methods, randomized smoothing, and more. For example, Metzen et al. \cite{metzen2017detecting}, use a detection method that works to augment current deep neural networks using a network called a “detector”, which works to root out any potential adversarial noise from the source by being trained to classify possible perturbations of an image. Another type of defense is called defensive distillation, as proposed in \cite{papernot2016limitations}. Defensive distillation is a technique which trains a second sub-model which is “distilled”, trained on soft labels predicted by its primary model. This type of defense is meant to work well on small, often unseen perturbations, and smooths the decision boundaries of an image. 

Other defenses include randomized smoothing, well executed in \cite{lecuyer2019certified}, a method that works by adding random noise to an input image and then work on classifying the smoothed version. Its goal, as inferred from the name, is to provide a smoothed prediction by averaging out over multiple noisy inputs, so it can be resistant to noise perturbations. This is one of the few methods that provide a probabilistic guarantee for some level of robustness. Xu et al., introduced feature Squeezing, a method which reduces the possible space for noise by literally squeezing the features of an image, using either spacial smoothing or bit-depth reduction \cite{xu2017feature}. The goal is to make it harder for manipulations to be made to the models predictions, since there is less space to create such manipulations.

More recent developments in defending against adversarial noise attacks in AI have led to several innovative strategies aimed at improving the robustness of deep neural networks \cite {liao2018defense}. One approach is adversarial augmentation, where defensive perturbations are preemptively applied to input data to ensure that any adversarial attack fails \cite{gu2014towards}. Another promising method is parametric noise injection, which involves adding trainable Gaussian noise at different layers of a DNN \cite{he2019parametric}. This noise is optimized during training to improve the model's resistance to black box attacks, making the network more resilient without sacrificing accuracy. These are only a few of the many different strategies that currently exist to catch adversarial noise, but what makes these attacks so dangerous is that they come in many different shapes, and any good defense against them requires a multi-faceted response in order to be successful in all cases.

Although there are multiple methods currently available to defend against adversarial attacks, most are not robust enough or fail to defend in certain cases. Research such as that done in Goodfellow et al. \cite{goodfellow2014explaining} has shown that even simple methods like FGSM can highlight the significant vulnerabilities of modern deep neural networks. FGSM uses gradient information in order to misrepresent linear approximations of loss for input points. It is an attack that is simple to create, not taxing to perform, and oftentimes extremely volatile. Even more recent state-of-the-art neural networks can be open to attack, as shown in \cite{carlini2017towards}, which used three separate attacks to effectively and consistently fool a model, both with and without its built in defenses. Athalye et al., further dives into the point that defenses can sometimes mean little to certain attacks, and concluded that many of the current so called robust defenses could easily be bypassed by the right type of adversarial attack \cite{athalye2018obfuscated}. 

Given the limitations discussed above and the critical importance of addressing adversarial attacks, we performed various analyses to explore potential solutions. Our contribution is aimed at better understanding the pattern of adversarial noise, how it functions, and its impact on classification performance. In this paper, we hypothesize that by performing pixel-wise analysis of noisy images, training the model on edge features, and combining edges with both noisy and clean images during retraining, we can gain a deeper understanding of the characteristics of adversarial noise, allowing the model to learn better for improved classification.

The rest of the paper organized as follows: Section 2 is about adversarial noise and how this type of noise alter edge pixels. In Section 3 we discussed the methods used in this study. Section 4 is the experimental results and discussion. In this section, several experiments have been performed to examine the impact of adversarial noise on image classification and how to improve the recognition accuracy by considering the image edges. Finally, we conclude in Section 5.

\section{Adversarial Noise}

    Adversarial noise in image classification are subtle and often unnoticeable perturbations introduced into an image that can cause a deep neural network, to misclassify the image. These changes are designed to be undetectable by the human eye, yet they can lead to a significantly degraded performance of a classifier. This noise exploits the vulnerabilities in the model’s decision boundary, leading to incorrect predictions.
    Adversarial examples are inputs modified by adversarial noise to produce bad outputs from machine learning models. These examples exploit the models' sensitivity to specific input alterations, revealing a disconnect between human perception and machine interpretation.

    There are many varying strategies when it comes to creating adversarial noise, each with different resource costs and affects on images. One of the important types of adversarial noise methods is the Fast Gradient Sign Method (FGSM), which aims to fool a network into misclassification through basic distortions in the image. As can be seen from Fig. 1, this noise can significantly affect the images by introducing some non important pixels, particularly around the edges and image boundaries. So, pixel-wise analyses may provide a better understanding of this type of noise and how to mitigate it.

\begin{figure}[h!]
    \centering
    \resizebox{\textwidth}{!}{%
    \begin{tabular}{ccc}
        \multicolumn{3}{c}{\makebox[\textwidth][c]{\hspace{1.3cm} Clean \hfill Clean Edges \hfill Noisy \hfill Noisy Edges}} \\
        
        % Label for the first dataset
        
        \rotatebox{90}{\shortstack[c]{\hspace{7mm} Images}} & 
        \includegraphics[width=0.5\textwidth]{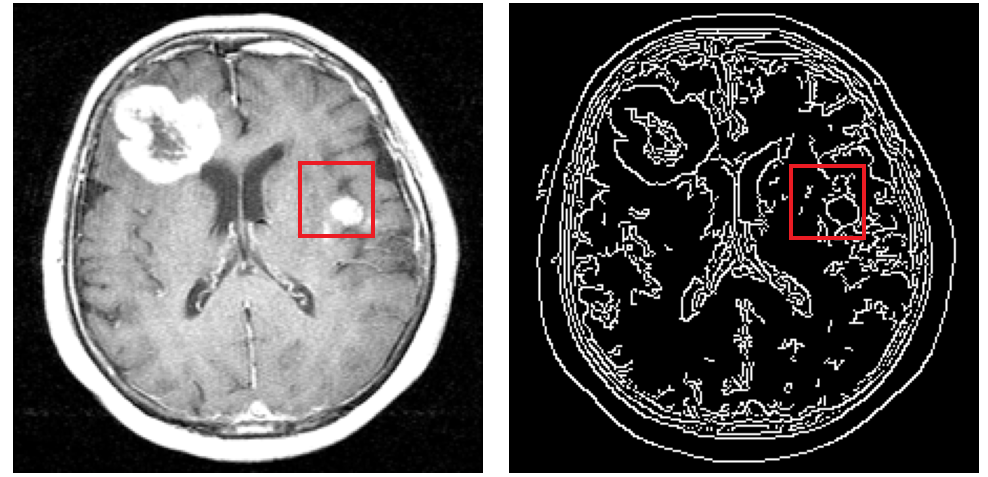} & 
        \hspace{-0.5cm} 
        \includegraphics[width=0.5\textwidth]{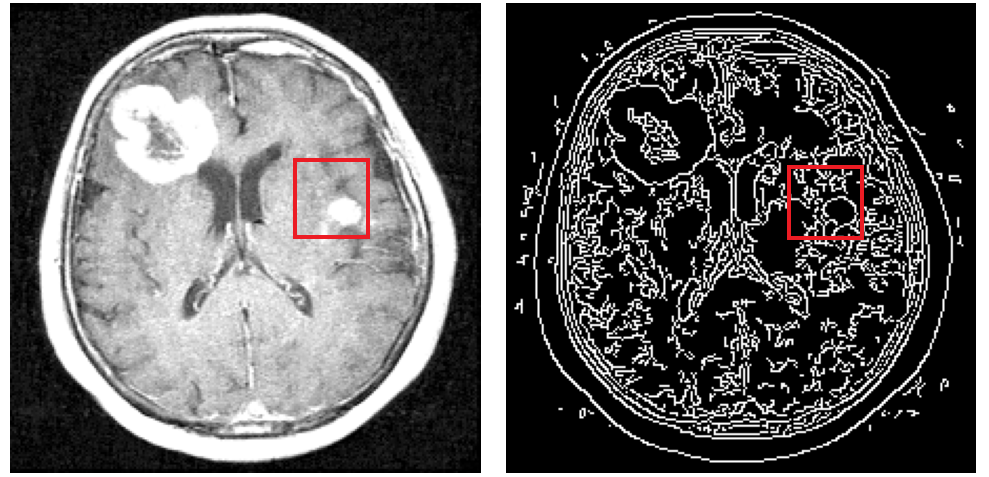} \\ 
        
        \rotatebox{90}{\shortstack[c]{\hspace{7mm} Pixels}} & 
        \includegraphics[width=0.5\textwidth]{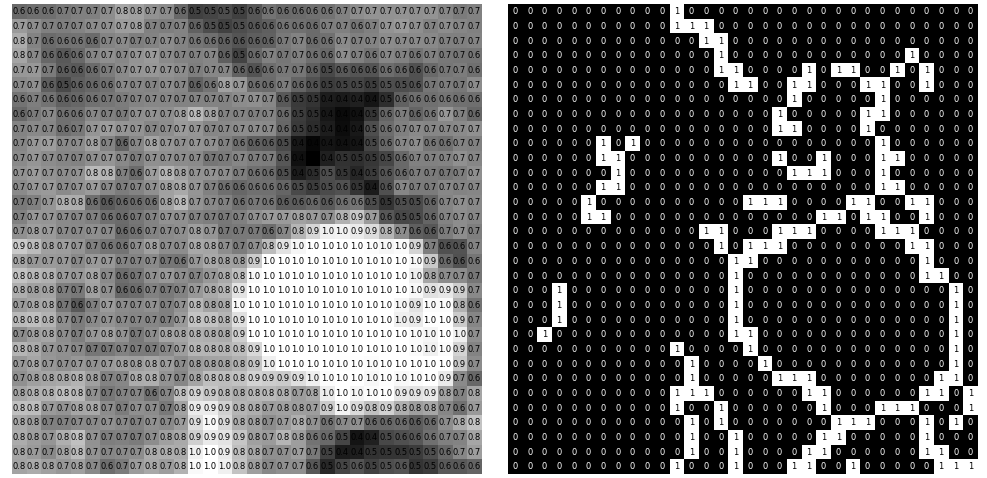} & 
        \hspace{-0.5cm} 
        \includegraphics[width=0.5\textwidth]{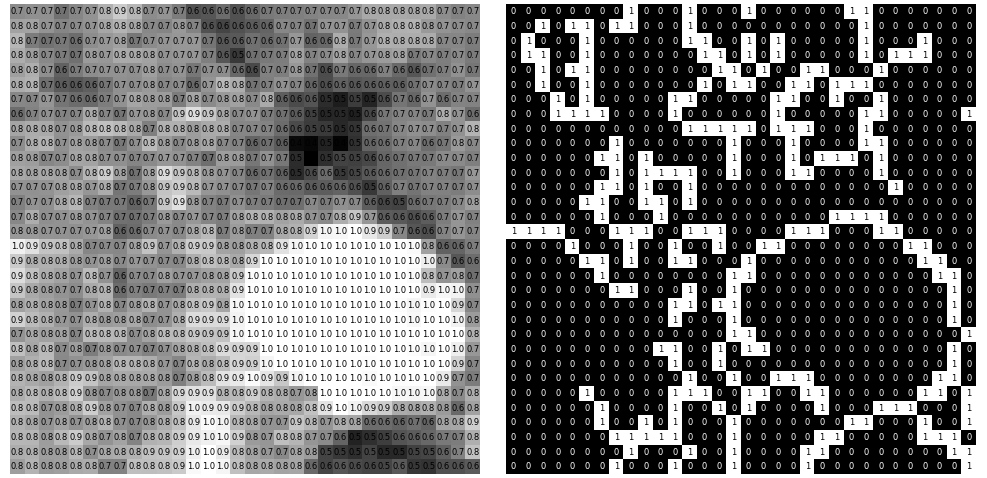} \\ 
        
        % Label for the second dataset
        \multicolumn{3}{l}{\hfill \textbf{(a)}} \hfill \\
        \vspace{0.5cm} \\
        \rotatebox{90}{\shortstack[c]{\hspace{7mm} Images}} & 
        \includegraphics[width=0.5\textwidth]{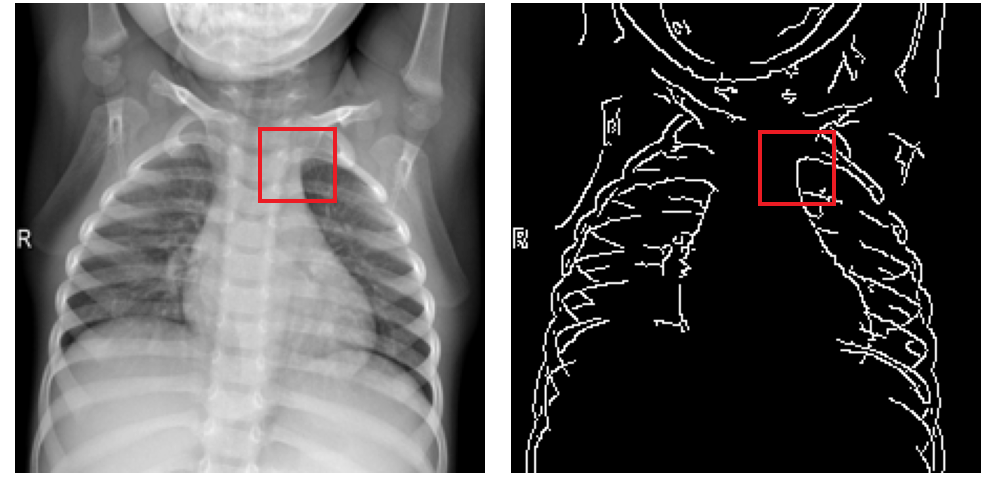} & 
        \hspace{-0.5cm} 
        \includegraphics[width=0.5\textwidth]{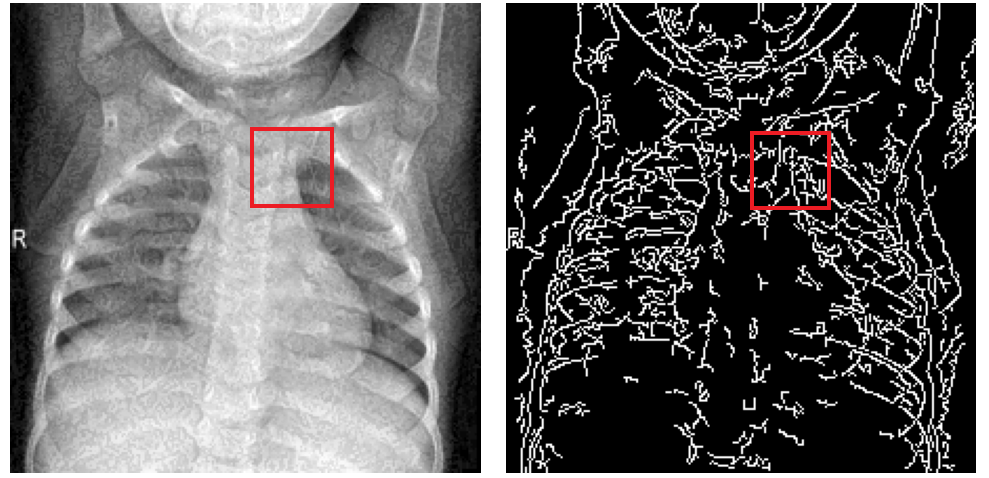} \\ 
        
        \rotatebox{90}{\shortstack[c]{\hspace{7mm} Pixels}} & 
        \includegraphics[width=0.5\textwidth]{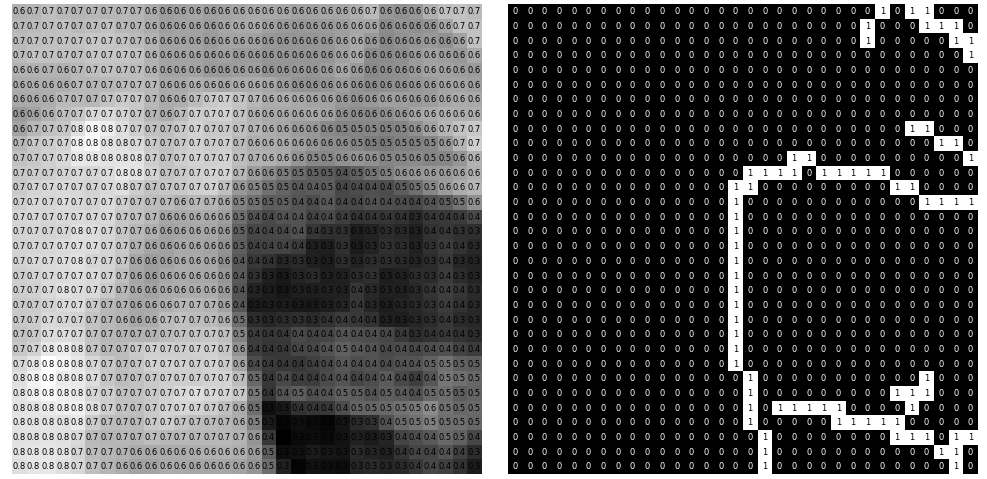} & 
        \hspace{-0.5cm} 
        \includegraphics[width=0.5\textwidth]{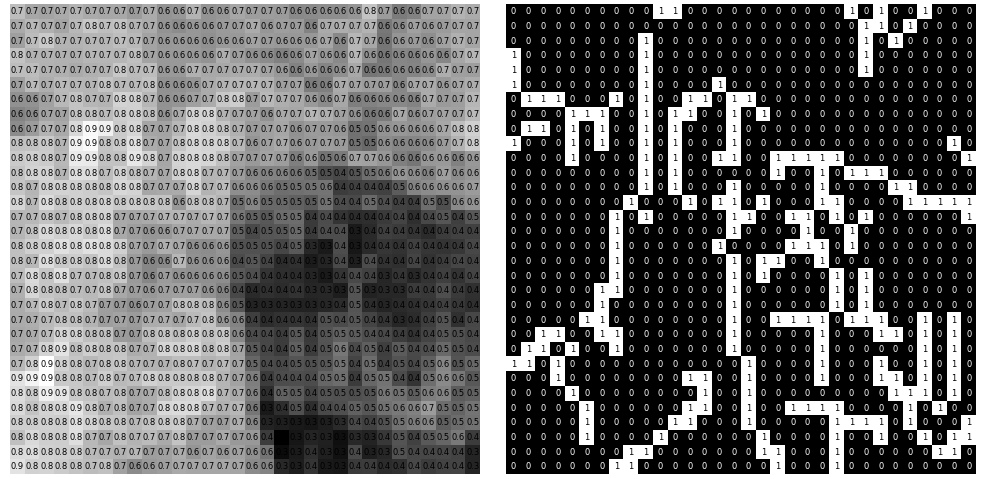} \\ 
        \multicolumn{3}{l}{\hfill \textbf{(b)}} \hfill \\
    \end{tabular}%
    }
    \caption{Visual comparison of clean and noisy images, along with their corresponding edges. Each patch is of size 32$\times$32, extracted from the original and noisy 224$\times$224 images, with noise added at an epsilon of 0.05. (a) illustrates examples from the Brain Tumor dataset \cite{hamada2020br35h}, and (b) shows samples from the COVID-19 dataset \cite{kermany2018large,cohen2020covid19imagedatacollection,chen2020mask,s7pw-jr18-20}.}

    \label{fig:example2}
\end{figure}
Adversarial attacks can be categorized based on the attacker's knowledge and objectives. Based on knowledge access, attacks are classified as white-box or black-box. In white-box attacks like the Fast Gradient Sign Method (FGSM) \cite{goodfellow2014explaining} and Projected Gradient Descent (PGD) \cite{geisler2024attacking}, the adversary has complete knowledge of the model architecture, parameters, and training data, allowing them to compute gradients and craft precise perturbations. FGSM uses the gradient of the loss with respect to the input image to create a perturbation that maximizes the loss, leading to misclassification \cite{goodfellow2014explaining}. PGD extends FGSM by applying multiple iterations of small adversarial steps, each time projecting the perturbed image back onto an epsilon-neighborhood around the original image to keep the perturbation within a specified bound. On the other hand, in black-box attacks, the attacker has no direct access to the model's internal parameters and relies solely on inputs and outputs. Examples generated for one model may successfully deceive another model, even if they have different architectures or were trained on different data. Attackers may train their own surrogate models to approximate the target model's behavior and generate adversarial examples based on the surrogate.

These attacks can be categorized as either targeted or untargeted, depending on the attack objectives. In targeted attacks, the adversary aims to misclassify the input into a specific, incorrect class, requiring crafting perturbations that not only cause misclassification but steer the output toward the desired target class. Untargeted attacks aim simply to cause the model to output any incorrect class, generally easier to execute as they require less precise manipulation.

%Perturbation measurement provides another categorization, dividing attacks into L0, L2, and L-infinity attacks. L0 attacks minimize the number of pixels altered, with the Jacobian-based Saliency Map Attack (JSMA) being an example that identifies the most influential pixels to change. L2 attacks focus on minimizing the overall energy of the perturbation; the attack by Carlini and Wagner (2017) generates perturbations with minimal L2 norm, making changes distributed but small \cite{carlini2017towards}. L-infinity attacks limit the maximum change to any pixel value, with FGSM and PGD falling into this category, constraining the perturbation so that each pixel alteration does not exceed a small epsilon value.

Adversarial noise also extends to the physical world, where attackers manipulate real-world objects to deceive models. Examples include adversarial patches, printed patterns or stickers added to objects that cause misclassification when captured by a camera, and 3D adversarial objects, which involve modifying the shape or texture of objects so they are misidentified by recognition systems \cite{brown2017adversarial}. These physical attacks demonstrate that adversarial vulnerabilities can have tangible, real-world consequences, affecting systems like facial recognition or autonomous vehicles.

The existence of adversarial noise presents significant challenges.  Security risks are paramount—in applications like autonomous driving, healthcare diagnostics, and security surveillance, adversarial attacks could lead to catastrophic consequences. Moreover, the ease with which adversarial examples can be generated calls into question the robustness of even the most advanced models, affecting their reliability. Designing defenses is complicated by the adaptive nature of attacks; many defense mechanisms are circumvented shortly after their introduction as attackers develop new methods, highlighting the complexity of achieving comprehensive defense \cite{madry2017towards}.

Addressing adversarial noise requires a multifaceted approach. Adversarial training incorporates adversarial examples into the training process to improve model robustness. Defensive distillation, proposed by Papernot et al. \cite{papernot2016limitations}, reduces model sensitivity by smoothing the decision boundaries. Randomized methods like randomized smoothing provide probabilistic guarantees of robustness by averaging predictions over noise-added inputs \cite{lecuyer2019certified}. Feature squeezing reduces the input space complexity to limit the attack surface \cite{xu2017feature}. Despite these efforts, achieving comprehensive defense remains an open challenge. The adaptive nature of adversarial attacks means that defenses must continually evolve.

\iffalse
\begin{figure}[h!]
    \centering
        \hspace{0.8cm}{\footnotesize Clean Images\hspace{1mm} Clean Edges \hspace{1mm}Noisy Images  Noisy Edges}\\
    \begin{tabular}{ll}
        \rotatebox{90}{\shortstack[c]{\tiny MRI Image 1 \\ \tiny Tumorous}} & \includegraphics[width=0.6\textwidth]{Edges/output-eight.png} \\
        \rotatebox{90}{\shortstack[c]{\tiny MRI Image 2 \\ \tiny Tumorous}} & \includegraphics[width=0.6\textwidth]{Edges/first-img.png} \\
        \rotatebox{90}{\shortstack[c]{\tiny MRI Image 3 \\ \tiny Tumorous}} & \includegraphics[width=0.6\textwidth]{Edges/four-img.png} \\
        \rotatebox{90}{\shortstack[c]{\tiny MRI Image 4 \\ \tiny Non-Tumorous}} & \includegraphics[width=0.6\textwidth]{Edges/output.png} \\
        \rotatebox{90}{\shortstack[c]{\tiny MRI Image 5 \\ \tiny Tumorous}} & \includegraphics[width=0.6\textwidth]{Edges/output-zero.png} \\
        \rotatebox{90}{\shortstack[c]{\tiny MRI Image 6 \\ \tiny Tumorous}} & \includegraphics[width=0.6\textwidth]{Edges/output-thirtythree.png} \\
        \rotatebox{90}{\shortstack[c]{\tiny MRI Image 7 \\ \tiny Tumorous}} & \includegraphics[width=0.6\textwidth]{Edges/output-three.png} \\
        \rotatebox{90}{\shortstack[c]{\tiny MRI Image 8 \\ \tiny Non-Tumorous}} & \includegraphics[width=0.6\textwidth]{Edges/output-eleven.png} \\
    \end{tabular}
    \caption{Brain MRI Images with Corresponding Adversarially Perturbed Images and their Edge Highlights}
    \label{fig:example}
\end{figure}
\fi

\section{Methodology}
%This section discusses how a specific type of adversarial noise affects classification models, leading to misclassification. We will evaluate conventional CNNs \cite{fukushima1980neocognitron}, ResNet50 \cite{he2016deep}, VGG16, and VGG19 \cite{simonyan2014very}, assessing their performance against this noise and evaluating the effects on the other models. This analysis helps us understand how white-box adversarial attacks impact both the targeted model and others.

\subsection{Fast Gradient Sign Method}
The adversarial noise in this paper is generated using the Fast Gradient Sign Method (FGSM) \cite{goodfellow2014explaining}, which as mentioned above, is a fast and easy strategy to generate adversaries by exploiting the vulnerabilities of deep learning models, particularly their sensitivity to input perturbations in high-dimensional spaces and linearity. FGSM generates minimal perturbations and targeted or untargeted alterations to the input data that can significantly throw models on classification tasks. This method calculates perturbations designed to maximize the model's error, thereby generating inputs that cause the model to misclassify. Let $\boldsymbol{\theta}$ represent the parameters of a model, $\mathbf{x}$ is the input, $y$ the true target label, and $J(\boldsymbol{\theta}, \mathbf{x}, y)$ the cost function used during training. The adversarial perturbation $\boldsymbol{\eta}$ is calculated as:
\begin{equation}
\boldsymbol{\eta} = \epsilon \, \text{sign}\left( \nabla_{\mathbf{x}} J(\boldsymbol{\theta}, \mathbf{x}, y) \right)
\end{equation}

Here, $\nabla_{\mathbf{x}} J(\boldsymbol{\theta}, \mathbf{x}, y)$ is the gradient of the cost function with respect to the input $\mathbf{x}$, pointing in the direction of the steepest ascent in the error. The $\text{sign}()$ function returns the sign of each gradient element, making a vector with components $+1$, $-1$, or $0$ that indicates the direction of change for each input feature. The scalar $\epsilon$ controls the magnitude of the perturbation, ensuring that it remains small and within the $\ell_{\infty}$-bounded constraint.

The adversarial example $\tilde{\mathbf{x}}$ is then generated by adding the perturbation $\boldsymbol{\eta}$ to the original input $\mathbf{x}$:
\begin{equation}
\tilde{\mathbf{x}} = \mathbf{x} + \boldsymbol{\eta}
\end{equation}

This perturbation process is computationally efficient and exploits the high-dimensional structure of the input space, where small, distributed changes across many features can collectively lead to a significant change in the model's output. 

\subsection{Canny Edge Detection}
A Canny image is a simplified, binary representation of an original image where the primary edges—boundaries where intensity sharply change—are highlighted. In this process, each image is first converted to grayscale and then transformed to an 8-bit format, ensuring compatibility with the Canny algorithm \cite{canny1986computational}. The algorithm detects edges through a series of steps, including noise reduction, gradient calculation, non-maximum suppression, and double thresholding, resulting in a clean edge map. The edges are detected by computing intensity gradients using Sobel operators. Gradient's magnitude (\( G \)) is calculated as:
\begin{equation}
G = \sqrt{G_x^2 + G_y^2}
\end{equation}

where \( G_x \) and \( G_y \) are the gradients in the x and  y directions. 

Two threshold values are applied to distinguish strong and weak edges. The resulting binary edge map is normalized to the range \([0,1]\) for further processing. Finally, the single-channel edge map is converted to RGB by replicating edge information across all three channels, ensuring compatibility with applications like object detection, feature extraction, and machine learning models.
In Fig. 4, we demonstrate how the original images and their edge representation look under an adversarial attack on the brain tumor images. It's almost impossible to perceive any noise on the natural images; however, it is more clear in the edges to visually detect noise affecting the image to mislead the classification models.

\subsection{Retraining Process}
Once we have both the noisy and clean versions of the image, we can combine or concatenate them to create a new dataset exclusively for retraining. The evaluation will focus on the noise generated after retraining to consistently assess the model’s ability to re-learn, mitigate adversarial threats, and enhance its robustness against such attacks (see Fig. 2). In-depth explanations are also provided in sections 4.1.1 and 4.1.3.
\begin{figure}[h!]
  \centering
  \includegraphics[width=1.0\textwidth]{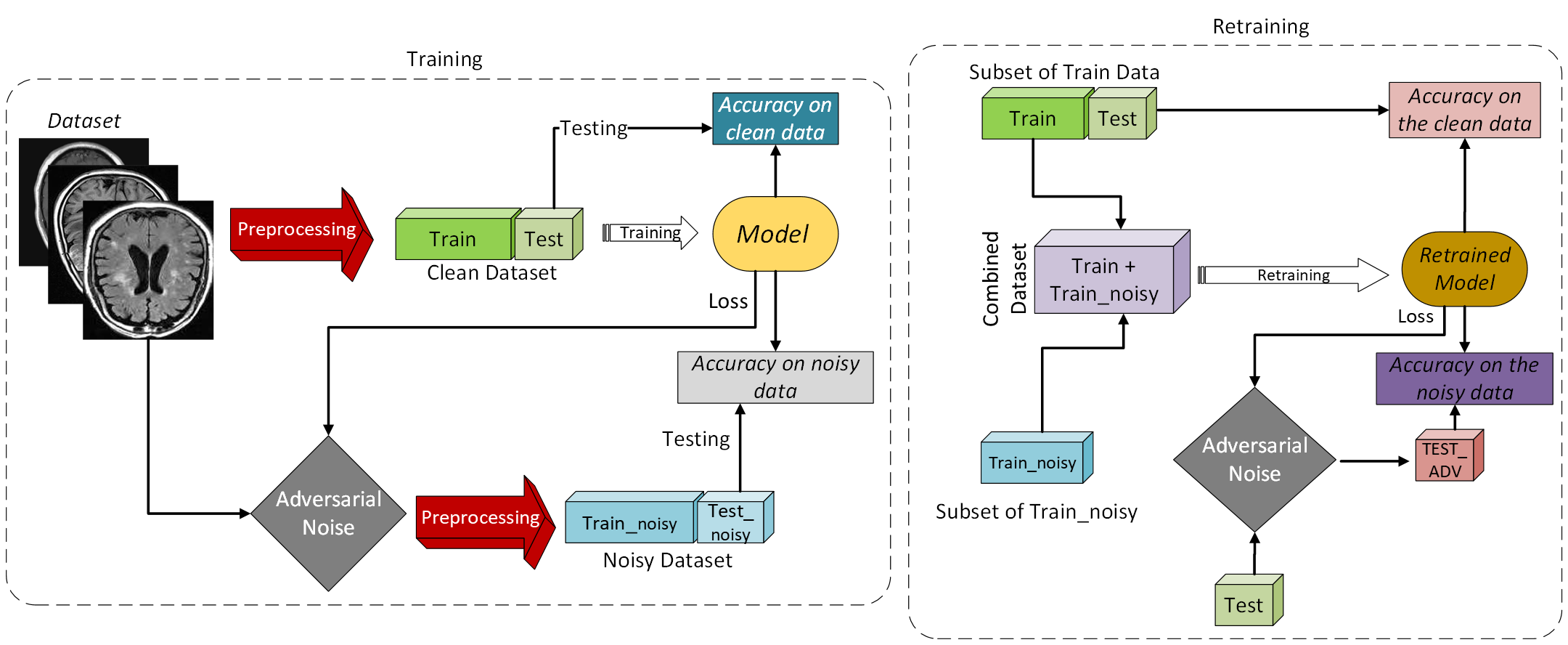}
  \caption{The process of training and retraining the model. Initially, the clean dataset is used to train and test a model, then a noisy dataset is created by introducing adversarial noise. The model is tested on both clean and noisy images to evaluate its baseline performance. A subset of the clean data is combined with the noisy training data, forming a new dataset for retraining. Finally, this retrained model undergoes evaluation against both clean and newly generated noisy images.}
  \label{fig:Pipeline}
\end{figure}
%\subsection{Deep Learning Models}

\section{Experimental results and discussions}
In this section, we performed several experiments to discuss and analyze the performance of various deep learning models when subjected to FGSM adversarial noise. We started by analyzing the impact of adversarial noise on several models. Then we discussed how to reduce the fooling rate by taking the edges into consideration. We now turn to the architectural details and designs of deep learning models for the following experiments, beginning with the Convolutional Neural Network (CNN).
CNN \cite{lecun1998gradient} architecture begins with an input layer designed for images of size 224 x 224 with 3 channels. It includes convolutional layers with 32, 64, and 128 filters, each followed by Batch-normalization, max-pooling, and concludes with a dense layer of 256 units and ReLU activation. A 50\% dropout layer is added to reduce overfitting, followed by an output layer with 2 units and soft-max activation for binary classification. The model is compiled using the Adam optimizer and categorical cross-entropy loss, with early stopping and learning rate reduction callbacks to enhance training efficiency. For transfer learning, we defined a base model (ResNet50 \cite{he2016deep}, InceptionV3 \cite{szegedy2016rethinking}, VGG16 and VGG19 \cite{simonyan2014very}, or DenseNet121\cite{huang2017densely}), initialized with ImageNet-pretrained weights. The input size was retained, and the top classification layers were removed. We also added a global average pooling layer, followed by a dropout, and a dense layer similar to the CNN. For training, the model was compiled using the Adam optimizer with the categorical cross-entropy loss function. The models were trained over 20 epochs with a batch size of 32, with two callback functions: EarlyStopping and ReduceLROnPlateau. Note that the visualizations in Figs. 3 and 4 were generated using CNN.  

\subsection{Experiment 1: Analysis on MRI Brain data}
\subsubsection{Impact analysis}
In this experiment, we used the  Brain Tumor dataset \cite{hamada2020br35h}, which consists tumorous (\texttt{yes}) and non-tumorous (\texttt{no}) categories. The images were preprocessed by resizing them to 224x224 pixels, normalizing pixel values, and assigning labels. The dataset was then split into training (80\%) and testing (20\%) sets. Adversarial noise was introduced to the test set to assess its impact on the performance of various models, including CNN, ResNet, VGG16, VGG19, InceptionV3, and DenseNet.
In this experiment, we aimed at analyzing the impact of adversarial noise on various deep learning-based models.

We started by training a classic Convolutional Neural Network (CNN) on clean images and then evaluated its performance on both clean and adversarially perturbed images, showing a significant drop in accuracy on noisy data (see Table 1). We then employed the same set of images to train and evaluate ResNet50. Unlike the classic CNN, the ResNet50's performance remained unaffected by the adversarial noise generated by the loss of the classic CNN. The model was compiled using the Adam optimizer and categorical cross-entropy loss and was trained with the same callback functions as the classic CNN. Next, we experimented with VGG16 and VGG19 models. Both models, known for their depth and simplicity, exhibited no significant change in performance when exposed to adversarial noise. The impact of adversarial noise on performance was minimal for both models. 

Additionally, Table 2 shows the fooling rates of various models when subjected to noise generated by each respective models. In other words, the rows of the table indicate the model used to generate the noise, while the columns show the fooling rates of the models when evaluated on that noise. The fooling rate, calculated using the equation below, represents the percentage of images that were incorrectly classified by each model when noise was added. 

\begin{equation}
F = \frac{N_{\text{fooled}}}{N_{\text{total}}}
\end{equation}

\noindent where $N_{\text{fooled}}$ is the number of samples for which the model's prediction changes after adversarial perturbation, and $N_{\text{total}}$ is the total number of samples.

The noise itself was generated by the model indicated in the leftmost column, and its effect on the models listed in the top row is shown in the corresponding cells. Additionally, noise generated from one model will only affect that specific model. For example, if the loss function used to generate adversarial examples is obtained from a CNN, this generated noise mainly affects the CNN and not the other models. We aim to improve the model's accuracy tested on noisy images derived from its own training. The following experiment is performed to see how the model can learn noisy patterns.

\begin{table}[h!]
\caption{Model Performance Comparison}
\vspace{-0.3cm}
\centering
\begin{tabular}{|c|c|c|c|c|c|}
\hline
\textbf{Model's Noise} & \textbf{Evaluated Model} & \multicolumn{2}{c|}{\textbf{Clean}} & \multicolumn{2}{c|}{\textbf{Noisy}} \\
\cline{3-6}
 & & Loss & Accuracy & Loss & Accuracy \\
\hline
\multirow{4}{*}{\textbf{CNN}} 
 & CNN & 0.15 & 0.97 & 4.95 & 0.50 \\ \cline{2-6} 
 & ResNet50 & 0.47 & 0.78 & 0.48 & 0.78 \\ \cline{2-6} 
 & VGG16 & 0.38 & 0.84 & 0.40 & 0.83 \\ \cline{2-6} 
 & VGG19 & 0.41 & 0.85 & 0.41 & 0.84 \\ \cline{2-6} 
 & InceptionV3 & 0.41 & 0.85 & 0.41 & 0.84 \\ \cline{2-6} 
 & DenseNet & 0.17 & 0.93 & 0.22 & 0.92 \\
\hline
\multirow{4}{*}{\textbf{ResNet50}} 
 & CNN & 0.15 & 0.95 & 0.13 & 0.95 \\ \cline{2-6} 
 & ResNet50 & 0.47 & 0.79 & 0.66 & 0.60 \\ \cline{2-6} 
 & VGG16 & 0.38 & 0.84 & 0.41 & 0.82 \\ \cline{2-6} 
 & VGG19 & 0.40 & 0.85 & 0.41 & 0.82  \\ \cline{2-6} 
 & InceptionV3 & 0.13 & 0.94 & 0.18 & 0.94 \\ \cline{2-6} 
 & DenseNet & 0.17 & 0.94 & 0.27 & 0.88 \\ 
\hline
\multirow{4}{*}{\textbf{VGG16}} 
 & CNN & 0.10 & 0.97 & 0.11 & 0.97 \\ \cline{2-6} 
 & ResNet50 & 0.47 & 0.79 & 0.47 & 0.79 \\ \cline{2-6} 
 & VGG16 & 0.38 & 0.85 & 1.81 & 0.19 \\ \cline{2-6} 
 & VGG19 & 0.40 & 0.85 & 0.69 & 0.59 \\ \cline{2-6} 
 & InceptionV3 & 0.12 & 0.96 & 0.49 & 0.78 \\ \cline{2-6} 
 & DenseNet & 0.17 & 0.93 & 0.32 & 0.85 \\
\hline
\multirow{4}{*}{\textbf{VGG19}} 
 & CNN & 0.10 & 0.95 & 0.10 & 0.95 \\ \cline{2-6} 
 & ResNet50 & 0.46 & 0.80 & 0.47 & 0.80 \\  \cline{2-6} 
 & VGG16 & 0.38 & 0.86 & 0.68 & 0.60 \\ \cline{2-6} 
 & VGG19 & 0.40 & 0.85 & 1.83 & 0.17 \\ \cline{2-6} 
 & InceptionV3 & 0.12 & 0.95 & 0.58 & 0.74 \\ \cline{2-6} 
 & DenseNet & 0.17 & 0.94 & 0.40 & 0.82 \\
\hline
\multirow{5}{*}{\textbf{InceptionV3}} 
 & CNN & 0.16 & 0.98 & 0.16 & 0.98 \\ \cline{2-6} 
 & ResNet50 & 0.48 & 0.78 & 0.48 & 0.78 \\  \cline{2-6} 
 & VGG16 & 0.39 & 0.84 & 0.51 & 0.75 \\ \cline{2-6} 
 & VGG19 & 0.41 & 0.85 & 0.84 & 0.55 \\ \cline{2-6} 
 & InceptionV3 & 0.11 & 0.95 & 4.44 & 0.10 \\ \cline{2-6} 
 & DenseNet & 0.17 & 0.93 & 0.22 & 0.92 \\
\hline
\multirow{5}{*}{\textbf{DenseNet}} 
 & CNN & 0.16 & 0.98 & 0.16 & 0.98 \\ \cline{2-6} 
 & ResNet50 & 0.48 & 0.78 & 0.48 & 0.78 \\  \cline{2-6} 
 & VGG16 & 0.39 & 0.84 & 0.51 & 0.75 \\ \cline{2-6} 
 & VGG19 & 0.41 & 0.85 & 0.84 & 0.55 \\ \cline{2-6} 
 & InceptionV3 & 0.11 & 0.95 & 0.45 & 0.80 \\ \cline{2-6} 
 & DenseNet & 0.19 & 0.93 & 5.31 & 0.08 \\
\hline
\end{tabular}
\end{table}

\begin{table}[h!]
\caption{Fooling rate of various models when subjected to noise generated by different models. Rows indicate the noise generator and columns show the evaluated model.}
\vspace{-0.3cm}
\centering
\begin{tabular}{|l|c|c|c|c|c|c|}
\hline
\diagbox{\textbf{Noise}}{\textbf{Tested}} 
    & \textbf{CNN} 
    & \textbf{RsNt50} 
    & \textbf{VGG16} 
    & \textbf{VGG19} 
    & \textbf{IcptV3} 
    & \textbf{DnsNt} \\ \hline
\textbf{CNN}        & 47\%   & 0\%    & 1\%  & 1\%  & 1\%  & 1\%  \\ \hline
\textbf{ResNet50}   & 0\% & 19\%  & 2\% & 3\% & 0\% & 6\%  \\ \hline
\textbf{VGG16}      & 0\%  & 0\%  & 64\%  & 24\%  & 14\% & 8\%  \\ \hline
\textbf{VGG19}      & 0\%  & 0\%    & 26\%    & 68\%    & 21\% & 12\% \\ \hline
\textbf{InceptionV3} & 0\%    & 0\%    & 11\% & 30\% & 85\% & 1\%  \\ \hline
\textbf{DenseNet}   & 0\%    & 0\%    & 19\% & 35\% & 15\% & 85\% \\ \hline
\end{tabular}
\label{tab:fooling_rate_comparison}
\end{table}

\subsubsection{Edge-based learning}

In this experiment, we assess the robustness of deep learning models using both normal and canny images from the Brain Tumor dataset. We trained Model A, a Convolutional Neural Network (CNN) trained on clean images, and then evaluated its performance, showing 97\% accuracy. To assess the robustness of the model against adversarial attacks, perturbations are introduced using the Fast Gradient Sign Method (FGSM) with an epsilon value of 0.015. The accuracy decreased from 97\% on clean data to 56\% after adding the noise. Performance was evaluated on both clean and adversarially perturbed (noisy) images, showing a significant drop in accuracy on noisy data.

We then trained Model B, which is trained on edges,  under the same conditions as Model A, but instead of using clean images, we trained and tested it on the canny-edge version of the dataset. We applied Canny edge detection to the entire Brain Tumor dataset using threshold values of 100 and 200 and realized that Model B performed well, achieving 95\% accuracy. After introducing perturbations, the accuracy dropped only to 86\%. At the same epsilon value using FGSM, Model B showed greater resilience to adversarial noise than Model A, which was trained on clean images. This approach of incorporating edges into the training process significantly improved the model's robustness to adversarial perturbations. To generalize the method, we conducted the same experiment with other common classification models such as ResNet50, InceptionV3, VGG16, VGG19, or DenseNet121 and observed similar results (see Table 3). Note that in these experiments, Model A refers to training on the original images, while Model B refers to training on the corresponding edge maps.
\begin{table}[h!]
\centering
\caption{Performance of the models on the Brain Tumor dataset, evaluated on clean and adversarial test sets. }
\vspace{-0.3cm}

\label{tab:combined_performance1}
\begin{tabular}{l c c c c c c c c}
\toprule
\multirow{2}{*}{\textbf{Model}} & \multicolumn{4}{c}{\textbf{Training}} & \multicolumn{4}{c}{\textbf{Re-training}} \\
\cmidrule(lr){2-5} \cmidrule(lr){6-9}
 & \multicolumn{2}{c}{Original} & \multicolumn{2}{c}{Edges} & \multicolumn{2}{c}{Original} & \multicolumn{2}{c}{Edges} \\
\cmidrule(lr){2-3} \cmidrule(lr){4-5} \cmidrule(lr){6-7} \cmidrule(lr){8-9}
 & Clean & Noisy & Clean & Noisy & Clean & Noisy & Clean & Noisy \\
\midrule
CNN         & 97\% & 56\% & 95\% & 86\% & 98\% & 74\% & 95\% & 91\% \\
ResNet50    & 79\% & 67\% & 82\% & 66\% & 79\% & 70\% & 81\% & 72\% \\
VGG16       & 85\% & 22\% & 77\% & 73\% & 76\% & 30\% & 76\% & 73\% \\
VGG19       & 85\% & 20\% & 77\% & 68\% & 78\% & 30\% & 78\% & 69\% \\
InceptionV3 & 94\% & 11\% & 89\% & 29\% & 91\% & 19\% & 77\% & 38\% \\
Dense-Net   & 93\% & 8\%  & 84\% & 40\% & 79\% & 21\% & 80\% & 51\% \\
\bottomrule
\end{tabular}

\end{table}

\subsubsection{Impact of re-training}

After training multiple deep learning models—such as a classic CNN, ResNet50, VGG16, VGG19, DenseNet, and InceptionV3—and evaluating Model A and Model B on both original images and their corresponding edges for classifying images as “tumor” or “no tumor”, we found that Canny edges are more robust to adversarial noise compared to the original images. We extended the previous experiment by introducing re-training for both normal and edges to study the nature of model's robustness and resilience. For retraining, we combined the clean and noisy images in 1:1 ratio, using only a subset of original training set for both the normal and canny edge image versions. We kept the size of the re-training dataset consistent for a fair evaluation, essentially dividing the types of images in equal parts. For edge images only, we combine 800 clean edges and 800 noisy edges resulting a size of 1600. Similarly, for normal images, we combined both clean and noisy images, ensuring an equal split to maintain a total size of 1,600—800 images for each type. All models were retrained using the same hyperparameters and callback functions as those used in the initial training.

After retraining, we re-evaluated the performance of classification models A and B on “Normal/Original” and “Edges” images. Model A, trained on “Normal” images, demonstrated a better ability, in most cases, to recover accuracy on noisy or adversarial images compared to Model B, trained on “Edges” images. Specifically, the accuracy of the CNN model A increased from 56\% to 74\% with “Normal” images, reflecting a 18\% improvement. In contrast, Model B, trained on “Edges” images, showed only a 5\% improvement in accuracy (see Table 3). However, the accuracy at the edges was already 86\% before retraining and increased to 91\% afterward. Although the accuracy varied slightly between runs , the results remained close to the average.

\subsubsection{Visualizing noise effects in both original and edge representations}

\begin{figure}[h!]
  \centering
  \includegraphics[width=1.0\textwidth]{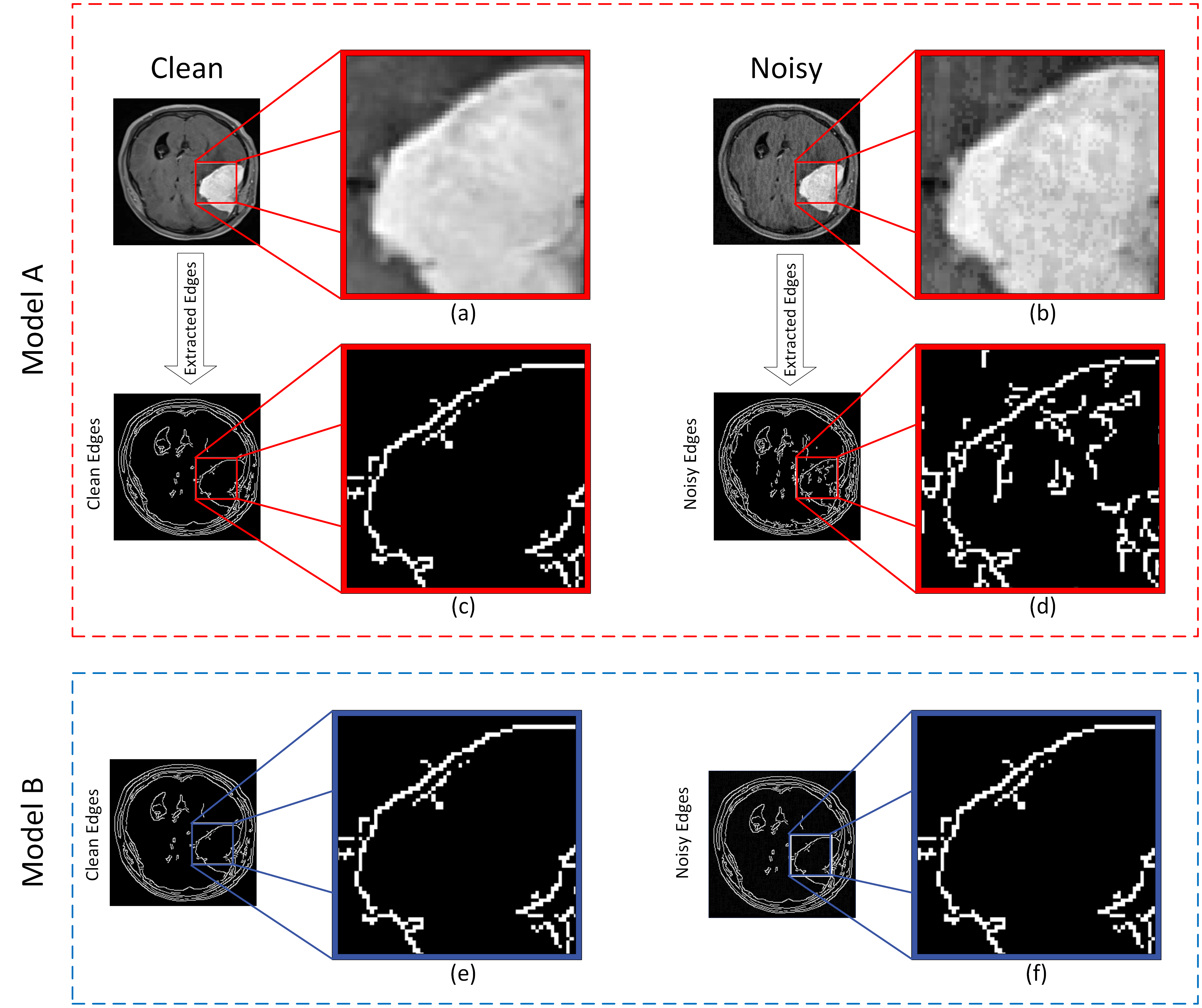}
  \caption{Visual comparison of noise generated by models A and B. (a) represents the original or clean image from the  Brain Tumor dataset, (b) represents the adversarial image generated using FGSM, (c) and (d) are the canny edge detected version of the (a) and (b) images, (e) and (f) represent the clean edges image and the corresponding adversarial image generated with Model B, respectively.}
  \label{fig:Pipeline2}
\end{figure}

In this experiment, as shown in Fig. 3, we explored how adversarial FGSM noise affects both original images and their corresponding edge representation for both Model A and Model B. Similar to the edge-training experiment, we first trained Model A on original images and then introduced small perturbations using FGSM adversarial noise with the $\epsilon$ value of 0.04. The noise is barely perceptible to the human eye when comparing the original or “clean" images to the corresponding “noisy" images. We then extracted Canny edges with threshold values of 100 and 200 and compared the images again. Noise becomes significantly more visible in edge-detected versions as seen in Fig. 4. In contrast, Model B was trained exclusively on a dataset that was pre-processed using Canny edge detection. After introducing FGSM noise with $\epsilon$ = 0.04, the resulting noise is barely perceptible, with no noticeable pixel differences for a similar epsilon value.

%\newpage
\subsection{Experiment 2: Analysis on COVID data}
\subsubsection{Data collection and preparation}

In this experiment, we used a COVID dataset \cite{kermany2018large,cohen2020covid19imagedatacollection,chen2020mask,s7pw-jr18-20} consisting of posteroanterior (PA) view chest X-ray images of normal, viral, and COVID-19 cases, totaling 1,823 images. The dataset was split into training and testing sets using an 80:20 ratio. Building on the approach adopted in the previous experiment, we evaluated the models’ robustness against adversarial examples generated from both original and edge-transformed images. We employed several well-known classification models, including CNN, ResNet, VGG16, VGG19, InceptionV3, and DenseNet. All models were retrained using the same hyperparameters, batch size of 32, and for 20 epochs, along with the callback functions used in the prior experiment.

\begin{figure}[h!]
  \centering
  \includegraphics[width=1.0\textwidth]{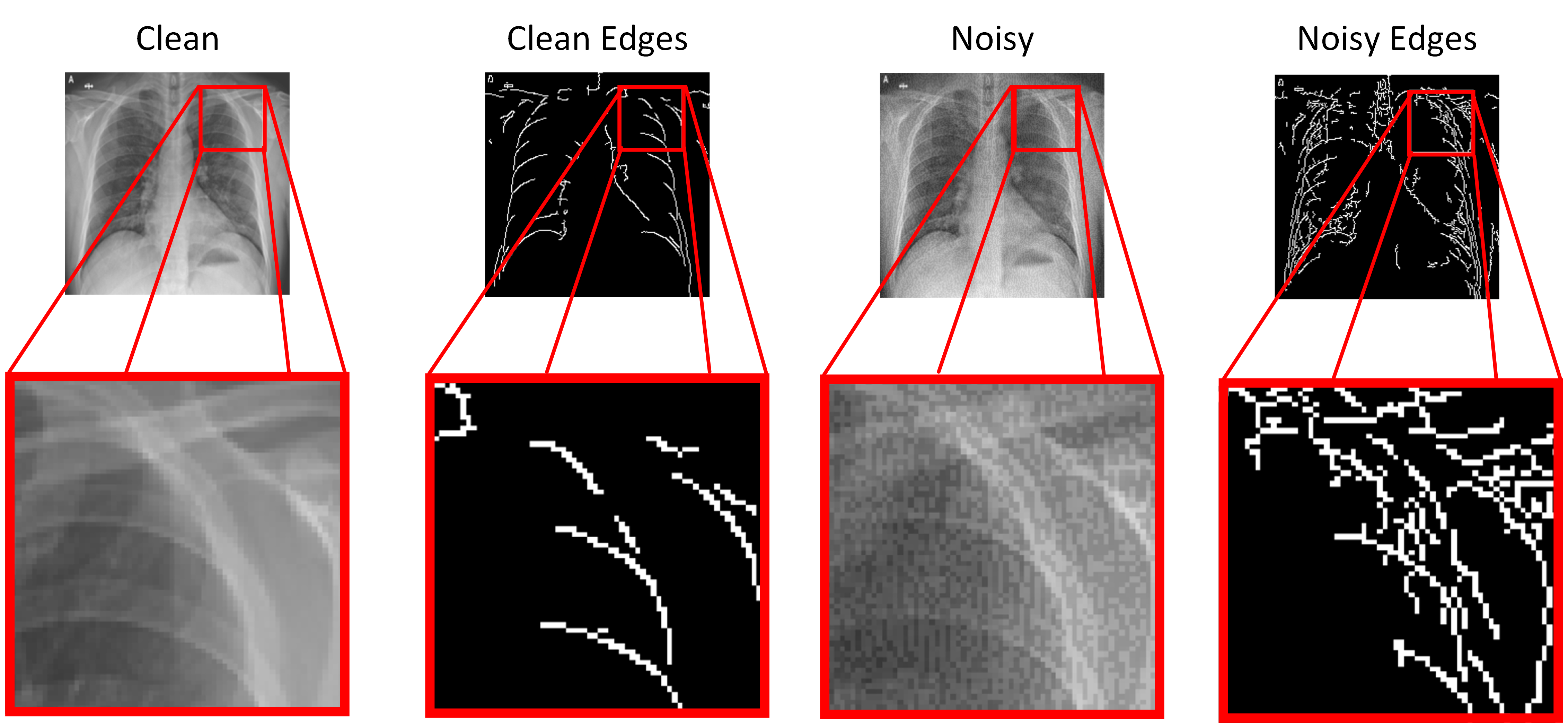}
  \caption{Visual comparison of noise generated for the COVID dataset.}
  \label{fig:Pipeline2}
\end{figure}

\subsubsection{Evaluation} 
We conducted multiple tests similar to the previous experiment, aiming to confirm the effectiveness of the edge-based training approach on this dataset and assess its generalization capability. We performed the pixel-wise analysis for COVID dataset as shown in Fig. 4. This analysis utilized both the clean and noisy versions of the images, along with their corresponding edge representations. After generating edges for both versions, we prepared a combined retraining dataset similar to a retraining experiment. We then evaluated the six models on the dataset to measure their accuracy and robustness, aiming to determine how original images and edges influenced each model's performance in terms of robustness (see Table 4).  The results remain consistent and show that training on edge images is more resilient against adversarial attacks generated using FGSM, compared to training on original images.

\begin{table}[h!]
\centering
\caption{Performance of various deep learning models on COVID, evaluated on clean and adversarial test sets.}
\vspace{-0.3cm}
\label{tab:combined_performance2}
\begin{tabular}{l c c c c c c c c}
\toprule
\multirow{2}{*}{\textbf{Model}} & \multicolumn{4}{c}{\textbf{Training}} & \multicolumn{4}{c}{\textbf{Re-training}} \\
\cmidrule(lr){2-5} \cmidrule(lr){6-9}
 & \multicolumn{2}{c}{Original} & \multicolumn{2}{c}{Edges} & \multicolumn{2}{c}{Original} & \multicolumn{2}{c}{Edges} \\
\cmidrule(lr){2-3} \cmidrule(lr){4-5} \cmidrule(lr){6-7} \cmidrule(lr){8-9}
 & Clean & Noisy & Clean & Noisy & Clean & Noisy & Clean & Noisy \\
\midrule
CNN         & 92\% & 68\% & 94\% & 91\% & 95\% & 76\% & 95\% & 92\% \\
ResNet50    & 75\% & 30\% & 87\% & 82\% & 75\% & 42\% & 88\% & 86\% \\
VGG16       & 89\% & 12\% & 86\% & 72\% & 81\% & 21\% & 88\% & 73\% \\
VGG19       & 87\% & 13\% & 87\% & 72\% & 84\% & 15\% & 85\% & 76\% \\
InceptionV3 & 95\% & 7\%  & 90\% & 55\% & 92\% & 29\% & 88\% & 67\% \\
Dense-Net   & 93\% & 8\%  & 86\% & 52\% & 89\% & 10\% & 91\% & 61\% \\

\bottomrule
\end{tabular}

\end{table}

\section{Conclusion}

Adversarial noise can deceive learning algorithms into making incorrect classifications by introducing small perturbations in images, thereby significantly affecting recognition accuracy. In this article, we analyze the impact of such noise on image classification. Our main hypothesis is that adversarial perturbations, which are designed to mislead deep learning algorithms, may be less effective against edge-based representations compared to raw images. To test this, we conducted pixel-wise analysis around image boundaries and edges to observe how adversarial noise alters pixel values in these regions. Several experiments were performed using brain tumor and COVID datasets to evaluate the performance of various deep learning models under adversarial conditions. The results suggest that edge-based features demonstrate significantly greater robustness to adversarial noise than raw image data, leading to improved recognition accuracy.

\bibliographystyle{splncs04}
\bibliography{bibliography}

\end{document}